\documentclass[letterpaper, 10 pt, conference]{ieeeconf}  

\IEEEoverridecommandlockouts                              

\usepackage{mathtools}
\usepackage{amsmath}
\usepackage{pdfpages}
\usepackage{cite}
\usepackage{hyperref}
\usepackage[normalem]{ulem}
\usepackage{amssymb}

\usepackage[ruled,vlined]{algorithm2e}

\title{\LARGE \bf
CLIPUNetr: Assisting Human-robot Interface for Uncalibrated Visual Servoing Control with CLIP-driven Referring Expression Segmentation
}

\author{Chen Jiang$^{\dagger}$, Yuchen Yang$^{\dagger}$ and Martin Jagersand$^{\dagger}$
\thanks{$^{\dagger}$Authors are with Department of Computing Science,
        University of Alberta, Edmonton AB., Canada, T6G 2E8.
        { 
           \tt\small \{cjiang2, yy17, mj7\}@ualberta.ca
        }
        }%
}

\begin{document}

\maketitle
\thispagestyle{empty}
\pagestyle{empty}

\begin{abstract}
The classical human-robot interface in uncalibrated image-based visual servoing (UIBVS) relies on either human annotations or semantic segmentation with categorical labels. Both methods fail to match natural human communication and convey rich semantics in manipulation tasks as effectively as natural language expressions. In this paper, we tackle this problem by using referring expression segmentation, which is a prompt-based approach, to provide more in-depth information for robot perception. To generate high-quality segmentation predictions from referring expressions, we propose CLIPUNetr - a new CLIP-driven referring expression segmentation network. CLIPUNetr leverages CLIP's strong vision-language representations to segment regions from referring expressions, while utilizing its ``U-shaped'' encoder-decoder architecture to generate predictions with sharper boundaries and finer structures. Furthermore, we propose a new pipeline to integrate CLIPUNetr into UIBVS and apply it to control robots in real-world environments. In experiments, our method improves boundary and structure measurements by an average of 120\% and can successfully assist real-world UIBVS control in an unstructured manipulation environment.


\end{abstract}

\section{Introduction}
Uncalibrated Image-Based Visual Servoing (UIBVS) \cite{gridseth2016vita} stands as a pivotal technique to cultivate position-based robot control. The pipeline of UIBVS can be broadly divided into two components: 1) perception, and 2) control. The perception phrase uses visual algorithms to analyze the environment from camera inputs, and to extract low-level image geometric features. The control phase takes in the geometric features and performs visuo-motor control. Originally, perception in UIBVS requires intense human labor for interfacing, in which the users have to review the scene in person and manually select the categories or shapes of the target \cite{gridseth2016vita}. More recent studies, integrated with AI\cite{myers2015affordance, do2018affordancenet, griffin2020video, xu2021affordance}, capture the semantics of the robot environment by training fix-class segmentation models. Though this process successfully reduces human involvement by interacting with specific categorical names in the human-robot interface, it still diverges from natural human communication. Predefined categories are inherently limited in encapsulating the rich, nuanced semantics of objects in manipulation tasks. As a result, they often fail to convey the depth of information that natural language expressions can offer.

\begin{figure}[t]
	\centering 
	\includegraphics[width=0.48\textwidth]{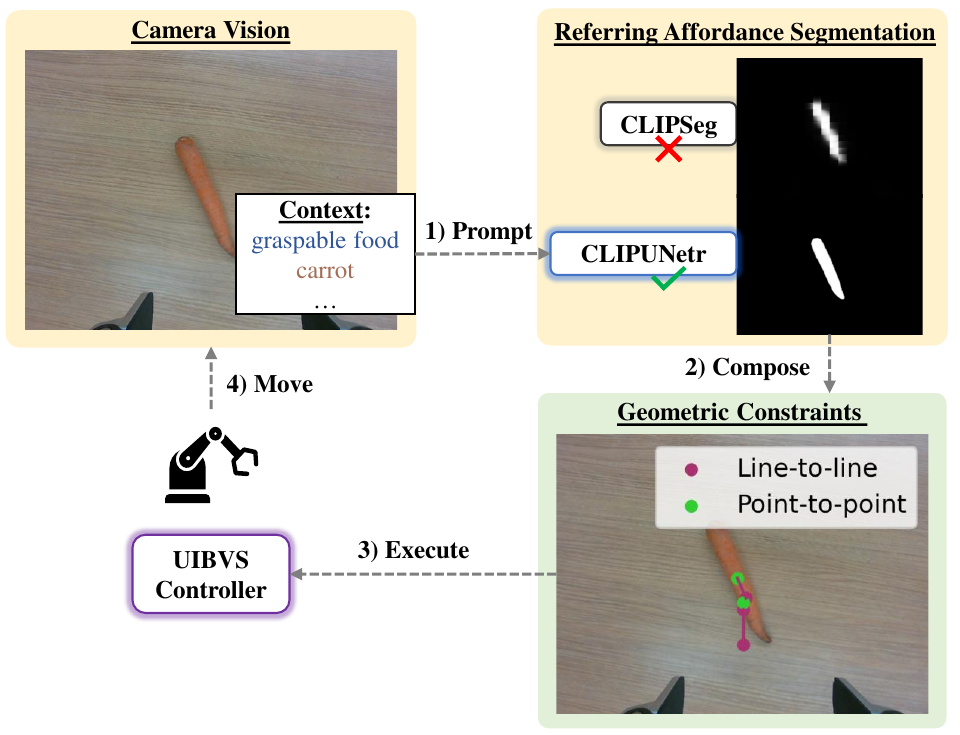}	
	\caption{Overview of our method. Given visual inputs, CLIPUNetr receives a referring expression, which is prompt-based, from a user to segment regions. Then, geometric constraints are composed from the prediction to execute UIBVS and move the robot.}
	\label{fig:teaser}%
\end{figure}

Recent developments of large-scale pretrained vision-language models (VLMs) like CLIP \cite{radford2021learning}, enabled strong vision-language representation trained from large-scale internet data. These advancements have given rise to methods such as CLIPSeg \cite{luddecke2022image}, capable of segmenting regions by descriptive texts. These methods have inspired us, highlighting a promising approach where the melding of rich language expression representation and image segmentation can potentially surpass the capabilities of traditional image-only semantic models, particularly in delineating more accurate and nuanced boundaries and structures. Motivated by this, we intend to further explore this hypothesis, aiming to develop a more intuitive and human-like communication approach by incorporating referring expression segmentation in the interfacing of UIBVS. 

In detail, we introduce CLIPUNetr\footnote{Our code is available at: \url{https://github.com/cjiang2/clipunetr}.}, a network adept at segmenting images into regions by leveraging user-specified referring expressions — sentence prompts that vividly describe a target object — alongside captured images. CLIPUNetr distinguishes itself by not only delivering precise segmentation results, marked by fine detailing in boundaries and structures, but also by mitigating the necessity for the laborious clicking traditionally associated with UIBVS interfaces and overcoming the limitations ingrained in category-specific approaches. Furthermore, we pioneer a new pipeline that seamlessly integrates CLIPUNetr as a perception module within the UIBVS framework. This incorporation amplifies the perception module's ability to adeptly handle affordances and other rich semantics during manipulation tasks, fostering a more intuitive and effective communication between users and the robot control system. A summary of our method can be seen in Figure \ref{fig:teaser}. We summarize our contributions as follows:

\begin{itemize}

\item We introduce a new referring expression based segmentation model, CLIPUNetr. The model is capable of capturing the multi-scale information through its ``U-shaped'' encoder-decoder, and utilizing the vision-language representations from CLIP.

\item We propose a new, modulated pipeline to perform UIBVS with prompt-based robot perception. We integrate CLIPUNetr as a perception module in this pipeline, enabling users to specify prompts for composing geometric motor skills.

\end{itemize}

\section{Related Work}
\subsection{Image Segmentation and Foundation Models} 
There have been various studies on image segmentation applied in various sub-fields, like salient object segmentation, affordance segmentation, etc. Models like BASNet \cite{qin2019basnet}, $U^2$Net \cite{qin2020u2} and DISNet \cite{qin2022highly} learn to encode multi-scale information. While the models achieve high spatial accuracy in boundary and structure, their predictions are fixed classes, and therefore restricted by categorical annotations. With the availability of pretrained models like vision transformers \cite{dosovitskiy2020image} and CLIP \cite{radford2021learning}, modern segmentation methods tend to use those large pretrained models as foundation models and finetune customized decoders on downstream tasks. Models like CLIPSeg \cite{luddecke2022image}, CRIS \cite{wang2022cris} and LSeg \cite{Li2022LanguagedrivenSS} leveraged image-text representations from CLIP for language-driven segmentation tasks, while models like SegViT \cite{zhang2022segvit} and UNETR \cite{hatamizadeh2022unetr} inferred attention masks from ViT with transformer decoders to generate segmentation results. However, evaluation only attends to regional accuracy (e.g. IoU), while spatial accuracy of boundary (e.g. S-measure \cite{fan2017structure}) is ignored.

\subsection{Robot Perception and Affordance}
One traditional way to construct robot perception is to train vision models and perform image segmentation to predict affordances for robot control. In Do et al \cite{do2018affordancenet} and Chu et al \cite{chu2019recognizing}, correct robot commands were inferred from affordances predicted by semantic models. K-VIL \cite{gao2023k} processed keypoints from demonstrations to form geometric constraints usable by imitation learning agents. Other methods \cite{manuelli2019kpam, xu2021affordance, gao2021kpam} combined affordance and keypoint predictions to guide robot control. With the introduction of language as control commands, methods like Nguyen et al \cite{nguyen2018translating} and Yang et al \cite{yang2019learning, yang2023watch} combined affordance prediction and vision-language models to construct modulated controllers that interface with language commands. However, training of the affordance segmentation models is expensive, and usually requires a large amount of annotated affordance data. Recently, foundation models like CLIP are utilized to train language-conditioned visuo-motor control policies. In this case, models are usually constructed as keypoint prediction modules like transporter \cite{kulkarni2019unsupervised}, and affordances are learned unsupervised from demonstrations. Typical models include CLIPort \cite{shridhar2022cliport} and Perceiver-actor \cite{shridhar2023perceiver}, where CLIP is used to encode actionable language commands, and affordances are represented as keypoints to guide the policy learning. However, the predicted affordances are less defined and structured versus the traditional approach.

\section{Methodology}
\subsection{Referring Expressions in Manipulation Workspace}
\begin{figure}[h!]
	\centering 
	\includegraphics[width=0.48\textwidth]{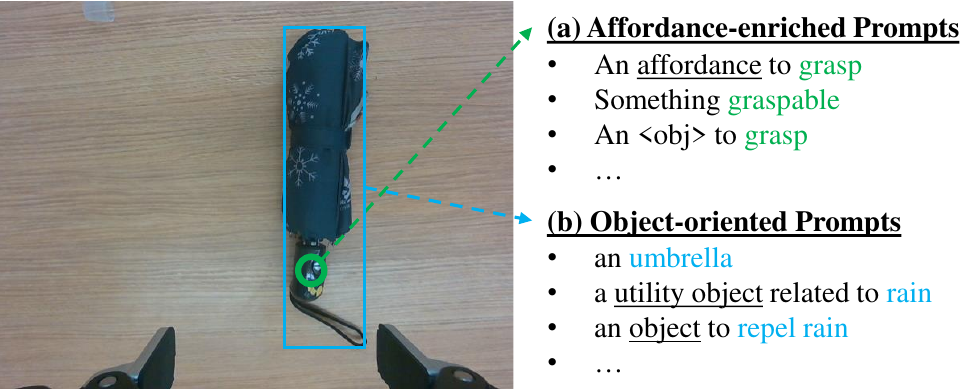}	
	\caption{Two types of prompts to refer to an umbrella object.} 
	\label{fig:prompt}%
\end{figure}

Given a referring expression as a prompt, and an image of a manipulation workspace with one or more objects, the goal is to segment the pixels that best match the expression. To refer to objects in the manipulation workspace in a contextual manner, the prompt describes important semantics like object naming, shape, functionality or affordances. Sample templates of prompts are visualized in Figure \ref{fig:prompt}, and categorized as follows:

\begin{itemize}

\item \textbf{Affordance-enriched Prompts} The prompt describes the functionality of an object, or parts of an object, in diverse, descriptive expressions. 

\item \textbf{Object-oriented Prompts} The prompt refers to an object explicitly by its common naming, or implicitly by describing shapes, colors, or intentional usages.

\end{itemize}

\noindent The two types of prompts can be combined to acquire prompts containing both affordance and object-oriented descriptions.

\subsection{Architecture}
\begin{figure*}[ht!]
\centering
\includegraphics[scale=0.66]{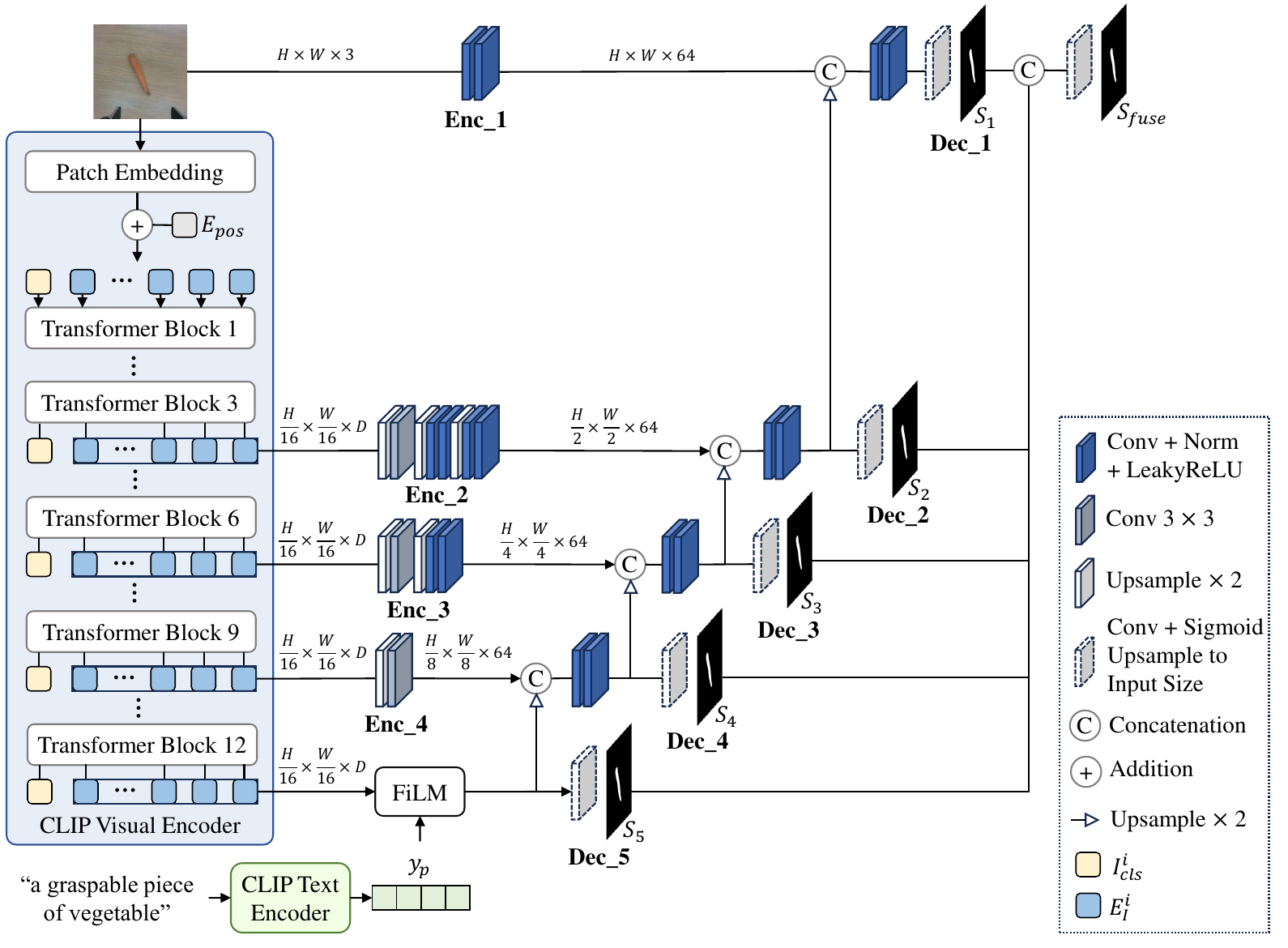}
\caption{Architecture of CLIPUNetr. Visual encoder encodes the image and generates token features and encoder features. The last token feature is conditioned with the text embeddings of the prompt through FiLM. The decoders take the conditioned token features and the encoder features to generate 5 side outputs, which are fused into a single segmentation probability map.}

\label{fig:clipunetr}
\end{figure*}

The proposed CLIPUNetr, visualized in Figure \ref{fig:clipunetr}, is a U-Net-like Encoder-Decoder network, which encodes the input image-prompt pairs, and decodes the segmentation probability maps.

\textbf{Visual Encoder}
First, CLIPUNetr infers CLIP's pretrained visual encoder. The encoder is a vision transformer (ViT) which takes an image input $I$ to capture semantic information. The image is first divided into $N$ patch embeddings $E_I$, where $N = HW/r^2$, $r$ is the patch resolution, $H$ and $W$ are the image height and width. A learnable class embedding $I_{cls}$ is concatenated with the patch embeddings, acquiring the token as $\{I_{cls}; E_I\}$, and the learnable positional embedding $E_{pos}$ is added to the token. Then, the token is processed by $L$ Transformer layers, composed by alternating multi-headed self-attention (MSA) and MLP blocks. The token from the final Transformer layer is then processed by an embedding head, outputting the image embedding $y_I$:

\begin{equation} \label{clip_image}
\begin{split}
E^j_{I} &= PatchEmbed(I^j), j=1,2,...,N \\
[I^0_{cls};E_0] &= [I_{cls};E_{I}] + E_{pos} \\
[I^{i+1}_{cls};E^{i+1}_{I}] &= {Layer}^i([I^i_{cls};E^i_{I}]), 0 \le i < L - 1 \\
y_I &= Head(I^{L}_{cls}) \\
\end{split}
\end{equation}

\noindent where ${Layer}^i$ denotes the $i^{th}$ transformer block. To support images with higher resolution, we interpolate $I_{cls}$ as discussed in Dosovitskiy et al \cite{dosovitskiy2020image}. 

Then, similar to UNETR \cite{hatamizadeh2022unetr}, additional encoders are attached, which take the same image input $I$, and tokens from the CLIP visual encoder. For image $I$, one block of $3 \times 3$ convolution, leaky ReLU and normalization layers are applied. For the token $\{I^{i}_{cls};E^{i}_I\}$, where $i \in \{3, 6, 9, 12\}$, the class embedding $I^{i}_{cls}$ is removed, and $E^{i}_I$ is reshaped from $N\times D$ to $r\times r\times D$. The reshaped embedding feature is then upsampled by a scale factor of 2, followed by consecutive $3 \times 3$ convolution, leaky ReLU and normalization layers for a number of times.

\textbf{Feature-wise Linear Modulation}
To inform the decoder with information from the prompt, the last encoder feature is conditioned with text embeddings through Feature-wise Linear Modulation (FiLM) \cite{dumoulin2018feature-wise}. Given a prompt $p$, CLIP text encoder encodes the prompt with transformer blocks, outputting a category [CLS] token $T_{cls}$. The token is projected by an embedding head to generate the text embedding $y_p$. For the token $E^L_I$ from $L^{th}$ layer, FiLM learns an affine transformation to perform conditional scaling:

\begin{equation} \label{film}
\begin{split}
FiLM(E^L_I) = \gamma (y_p) E^L_I + \beta (y_p) \\
\end{split}
\end{equation}

\noindent where $\gamma$ and $\beta$ are learnable parameters.

\textbf{Segmentation Decoder} 
To generate the resulting probability maps with high resolution, we enable feature scaling with side outputs in the construction of CLIPUNetr's U-shaped segmentation decoder, inspired by BASNet \cite{qin2019basnet}. First, the language-conditioned token $FiLM(E^L_I)$ is upsampled with a scale factor of 2, followed by consecutive 3 $\times$ 3 convolution, leaky ReLU and normalization layers. Then, the output is upsampled and merged with the feature of the previous transformer output via skip connections. The concatenated feature is again processed by another consecutive 3 $\times$ 3 convolution, leaky ReLU and normalization layers. The process is repeated till the output features are upsampled to the original input resolution, acquiring $5$ decoding features. Then, we produce side output probability maps, and refine the side outputs to generate the final probability map. Each decoder feature will be fed into a 3 $\times$ 3 convolution, followed with upsampling and a sigmoid function. The process generates 5 side output probability maps $S_5$, $S_4$, $S_3$, $S_2$, $S_1$. The side outputs are concatenated and refined by a 1$\times$1 convolution and a sigmoid function, generating the final probability map $S_{fuse}$.

\textbf{Hybrid Loss}
Similar to BASNet \cite{qin2019basnet}, we define loss as the summation of all outputs:

\begin{equation} \label{loss_sum}
\begin{split}
\mathcal{L} = {\alpha}_{fuse} {\ell}_{fuse} + \sum_{k=1}^{5} ({\alpha}^k_{side} {\ell}^k_{side})
\end{split}
\end{equation}

\noindent where ${\ell}_{fuse}$ is the loss for fused probability map, ${\ell}^k_{side}$ is the loss for side probability map, ${\alpha}_{fuse}$ and ${\alpha}^k_{side}$ are the weights of each loss term . Each individual loss term $\ell$ is calculated by a hybrid loss, which is composed by the summation of binary cross entropy, SSIM and IoU loss:

\begin{equation} \label{loss_mixture}
\begin{split}
{\ell} = {\ell}_{bce} + {\ell}_{ssim} + {\ell}_{iou}
\end{split}
\end{equation}

\noindent Utilizing a hybrid loss forces the decoder to focus on the foreground and preserve the structure of the predicted segmentation results, especially near the boundary.

\subsection{Geometric Constraint Composition}
To align the robot end effector position to a target position in the manipulation workspace, we construct geometric constraints \cite{gridseth2016vita} from image geometry, which specify the alignment context. In detail, given a list of ordered keypoints $\textbf{f}=\{f_1, f_2, ..., f_n\}$ in homogeneous coordinates, a geometric constraint $T$ is a task function that maps the keypoints into $\{0, 1\}$, $T(f)=0$ when the robot configuration is aligned with the task description, $T(f)=1$ when the alignment is violated (e.g. unparalleled lines). In the scope of this paper, four basic types of task descriptions are considered:

\begin{equation} \label{visual_tasks}
\begin{aligned}
T_{pp}(\textbf{f}) &= f_2 - f_1 \\
T_{pl}(\textbf{f}) &= f_1 \cdot l_{23} \\
T_{ll}(\textbf{f}) &= f_1 \cdot l_{34} + f_2 \cdot l_{34} \\
T_{par}(\textbf{f}) &= l_{12} \times l_{34} \\
\end{aligned}
\end{equation}

\noindent where $T_{pp}$, $T_{pl}$, $T_{ll}$, and $T_{par}$ are denoted as point-to-point (p2p), point-to-line (p2l), line-to-line (l2l), and parallel-line (par) task, respectively. A line $l_{ij}$ is denoted by the cross product of two points $f_i$ and $f_j$.

\begin{figure}[h!]
	\centering 
	\includegraphics[width=0.48\textwidth]{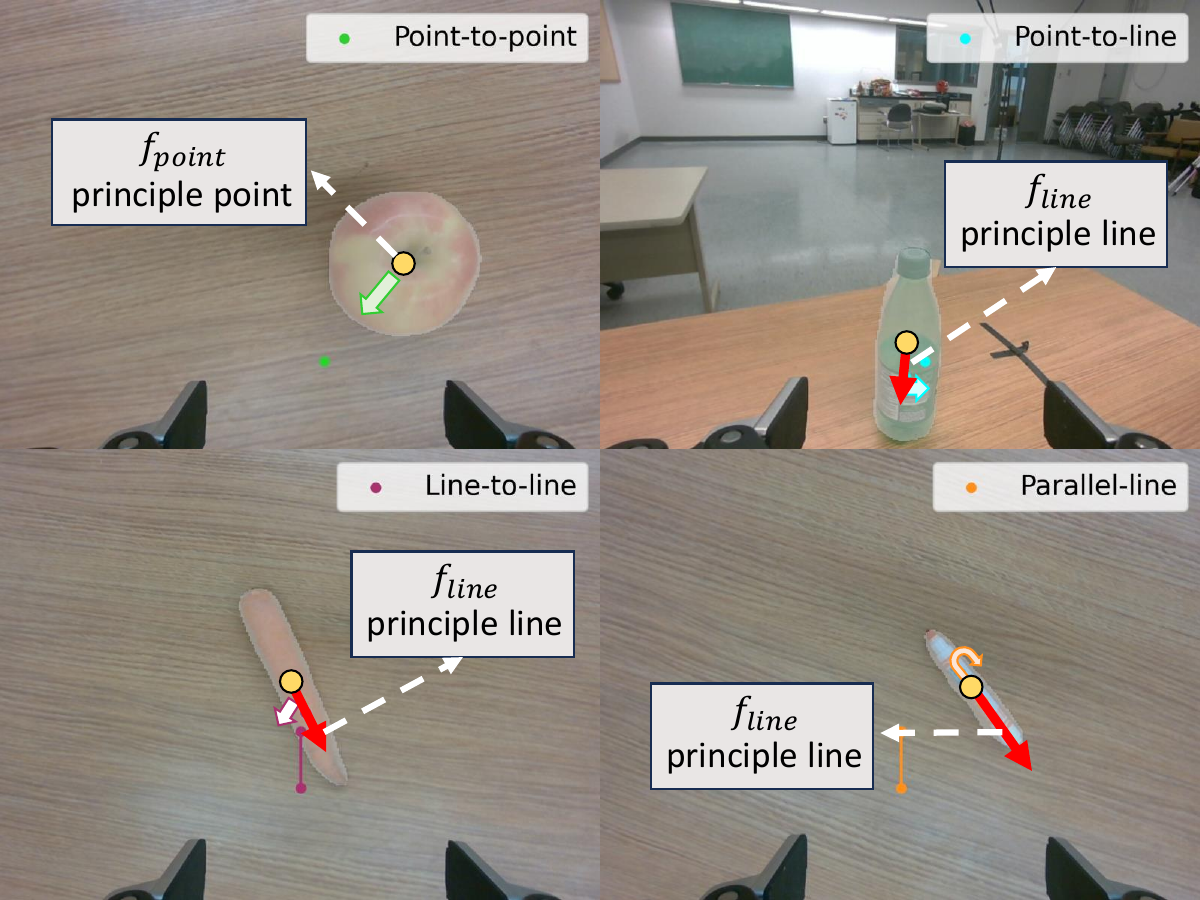}	
	\caption{Heuristics to compose four types of geometric constraints from the predictions of CLIPUNetr.} 
	\label{fig:pca}%
\end{figure}

Given the output probability map $M$ from CLIPUNetr and the prompt $p$, the process to compose geometric constraint in an eye-in-hand camera configuration is expanded as follows: First, the probability map is used as an image descriptor function $M(F)=(M(f_1), M(f_2), ..., M(f_k), ..., M(f_K))$, where $M(f_k)$ denotes the probability score of $p$ at a keypoint location $f_k$, $F=\{f_1, f_2, ..., f_k, ..., f_K\}$ defines the full image grid, where $K=HW$. Next, by filtering the probability score with a threshold of 0.5, we obtain a set of candidate keypoints $F_{candidates}$, where $f_k \in F_{candidates}$ and $M(f_k) > 0.5$. Then, PCA is used to analyze the variance of $F_{candidates}$, obtaining the principal point $f_{point}$ and principal lines $f_{line}$. Last, simple heuristics are used to compose the constraints, visualized in Figure \ref{fig:pca}. For a p2p task, the target position $f_2=f_{point}$ aligns to the end effector position $f_1$, heuristically set as $f_1=(W/2, 4H/5, 1)$. Similarly for a p2l task, the end effector position $f_1$ is set as $(W/2, 4H/5, 1)$, while the target line $l_{23}$ is set as $f_{line}$. For l2l and par task, the end effector orientation $l_{12}$ is defined as the vertical line passing through mid image center, while the target line $l_{34}$ is set as $f_{line}$.

\subsection{UIBVS with CLIPUNetr in Perception}
UIBVS \cite{jagersand1997experimental} defines the visual-motor control law, where a robot can be controlled to reach a desired joint configuration from image geometric inputs. The equation is denoted as:

\begin{equation} \label{vs_error}
\dot{e} = J_{u}(q)\dot{q}
\end{equation}

\noindent where $\dot{q}$ is the control input of a robot with $N$ joints, $J_{u}$ is the visuo-motor Jacobian. The values of the geometric constraints, calculated by combining heuristics and referring expression segmentation, are used as the error signal $\dot{e}$. Then, Broyden update is performed in replacement of camera calibration and analytical Jacobian calculation:

\begin{equation} \label{broyden_update}
\hat{J}^{(k+1)}_u = \hat{J}^{(k)}_u + \lambda \frac{(e - \hat{J}_u^{(k)} \Delta q ) \Delta q^T}{\Delta q^{T}  \Delta q + \epsilon}
\end{equation}

\noindent where $\lambda$ is the weight of the rank one Broyden update. 

Following the above definitions, Algorithm \ref{alg:perception} shows our pipeline to interface UIBVS with CLIPUNetr. Given an image observation $I_{t}$ at time $t$, the human operator first decides the geometric constraints to compose, and then specifies a prompt $p$ to initiate the robot perception. For each $I_{t}$, CLIPUNetr will be inferred, generating the probability map $M_t$. $m$ geometric constraints are composed following the discussed strategy, and are inputted into the UIBVS controller, generating the joint commands to move the robot. 

\begin{algorithm}[h!]
\SetAlgoLined
\textbf{Inputs:} A randomly sampled image frame $I_{t}$. Geometric constraints, text prompt $p$, specified by a human operator.\\
\KwResult{Joint command $\Delta J$.}
\While{True}{
    sample $I_t$\;
    $M_{t}$ = CLIPUNetr($I_{t}$, $p$)\;
    $f_{point}$, $f_{line}$ = PCA($M_{t}$)\;
    $\{T_1, ..., T_m\}$ = ComposeConstraint($f_{point}$, $f_{line}$)\;
    $\Delta J$ = UIBVS($\{T_1, ..., T_m\}$)\;
}
 \caption{Robot perception to compose geometric constraints and perform UIBVS control.}
 \label{alg:perception}
\end{algorithm}

\section{Experiments}
\subsection{Experimental Setup}
\textbf{Datasets}
We use the publicly available PhraseCut and UMD+GT datasets for experiments. PhraseCut dataset \cite{wu2020phrasecut} contains 340,000 phrases with associating regional referring expression prompts. UMD+GT dataset \cite{xu2021affordance} contains 30,000 RGBD images of 104 objects, originally annotated with 6 fix-class affordance labels. To generate referring expressions for UMD+GT dataset, we manually re-annotate the images based on 39 diverse prompt templates. In training, we randomly sample one of the prompts, and in testing, we replace the affordance label with one object-oriented prompt, and one affordance-enriched prompt. Prediction is done twice and we take the average metric scores of the two predictions. To evaluate object boundary and structure on both datasets, we use four metrics studied in salient object segmentation: Mean Absolute Error (MAE), Structure measure, weighted F-measure, and max F-measure. 

\textbf{Robot Control}
The robot control is evaluated offline and online. In offline setup, 19 robot manipulation tasks of moving and grasping, controlled by the classical manual clicking interface \cite{gridseth2016vita} and performed by a Kinova Gen3, are recorded using an eye-in-hand Intel Realsense D405 camera. Masks are annotated and evaluated every ${10}^{th}$ frame. 

In online setup, CLIPUNetr is integrated as a perception module to replace visual tracking with manual clicking. We compare the modified interface against the classical interface \cite{gridseth2016vita} by completing 12 robot manipulation tasks of moving and grasping with 5 food objects (apple, red pepper, lemon, carrot, banana), 2 utility objects (tennis ball, umbrella) and 4 marker pens. 3 attempts are allowed for the user to complete a task. The task is successful if the robot approaches and grasps the object without falling. We report the success rate for the three categories of tasks.

\textbf{Implementation Details}
The model is implemented using PyTorch, using the same hyperparameters from CLIPSeg \cite{luddecke2022image}. A batch size of 32 is used to train the models on a single Nvidia V100 GPU. For UIBVS control, joint 1, 2, 6 and 7 are used for table-top manipulation, and the implementation is done in ROS. A $\lambda$ of 0.05 is set for Broyden's update. Joint angle commands are published in 1 Hz to move the robot end effector in small amounts, until convergence.

\subsection{Results for Referring Expression Segmentation}

\begin{table}[h!]
\centering
\scalebox{0.97}{
\begin{tabular}{l c c c c c c c} 
 \hline
Dataset & Model & MAE$\downarrow$ & $S_m\uparrow$ & ${wF}_{\beta} \uparrow$ & $maxF_{\beta} \uparrow$ \\ 
 \hline
PhraseCut & CLIPUNetr & \textbf{0.0840} & 0.6975 & \textbf{0.5375} & \textbf{0.5887} \\
 & CLIPSeg \cite{dosovitskiy2020image} & 0.1343 & 0.6629 & 0.3836 & 0.5658 \\
 & CLIPSeg* & 0.1098 & 0.6865 & 0.4263 & 0.5840 \\
\hline
UMD+GT & CLIPUNetr & \textbf{0.0025} & \textbf{0.8968} & \textbf{0.7770} & 0.8021 \\
 & CLIPSeg \cite{dosovitskiy2020image} & 0.1714 & 0.5447 & 0.1165 & 0.3335 \\
 & CLIPSeg* & 0.0064 & 0.7858 & 0.4705 & 0.6027 \\
 & AffKp \cite{xu2021affordance} & 0.0044 & 0.8756 & 0.6753 & \textbf{0.8030} \\

\hline
\end{tabular}
}
\caption{Qualitative results for referring expression segmentation.}
\label{Table_qualitative}
\end{table}

\textbf{Qualitative Evaluation} Table \ref{Table_qualitative} shows the qualitative results of referring expression segmentation on PhraseCut and UMD+GT datasets. For fair comparison, an reimplementation of CLIPSeg is trained following our training regime (CLIPSeg*). In summary, CLIPUNetr is able to outperform CLIPSeg on both PhraseCut dataset and UMD+GT dataset, reflecting CLIPUNetr's capability of generating good quality predictions. For UMD+GT dataset, CLIPUNetr surpasses the performance of AffKp, a supervised network trained only with fix-class labels. This proves that leveraging image-text representations from CLIP is beneficial for learning.

\begin{figure}[h!]
	\centering 
	\includegraphics[width=0.48\textwidth]{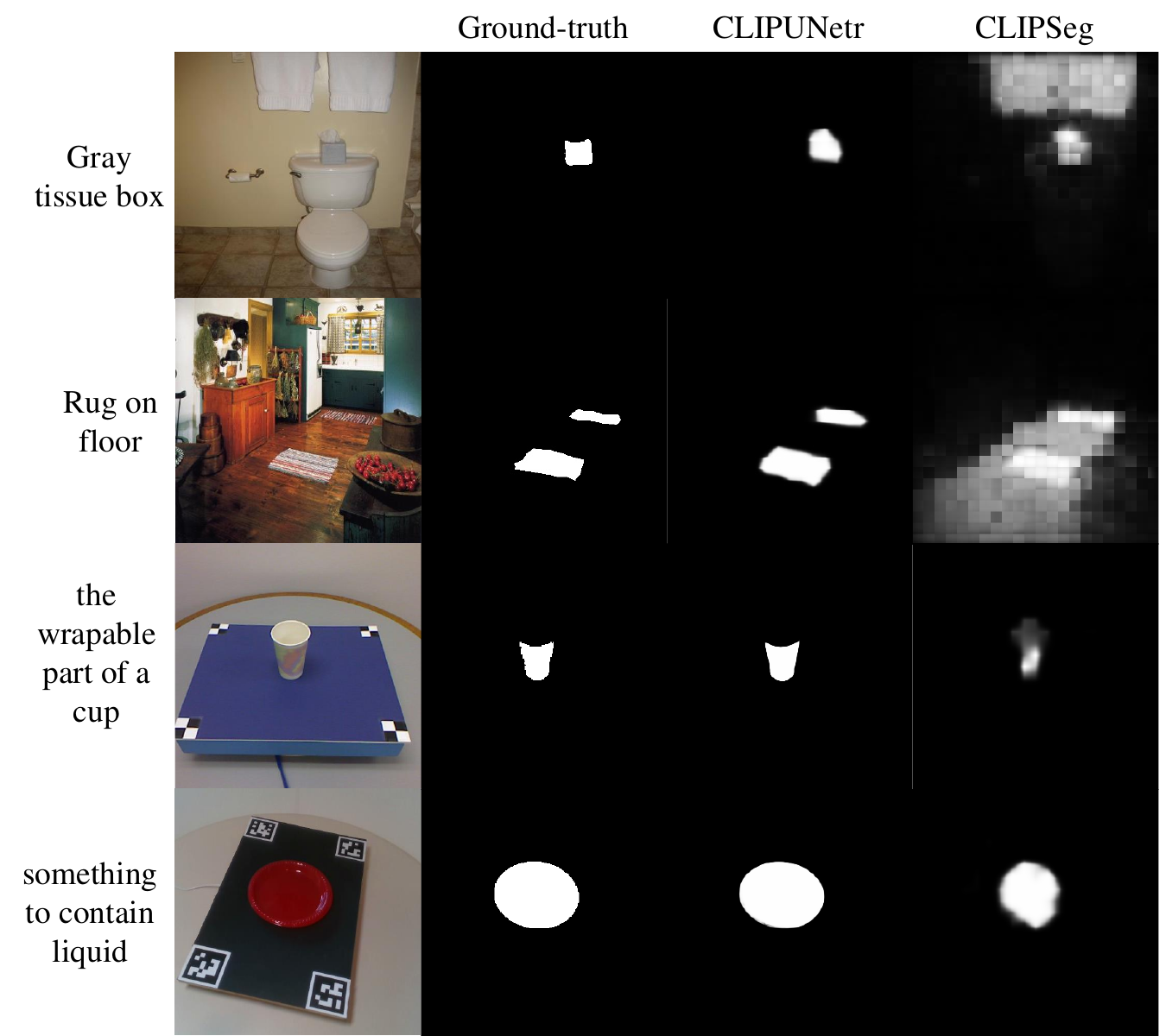}	
	\caption{Quantitative results on PhraseCut dataset and UMD+GT dataset.} 
	\label{fig:quantitative}%
\end{figure}

\textbf{Quantitative Evaluation} To further illustrate the superior performance of CLIPUNetr, Figure \ref{fig:quantitative} visualizes the prediction results from CLIPUNetr vs. CLIPSeg. As shown, predictions from CLIPSeg contain checkerboard artifacts, caused by abruptly upsampling low-resolution decoding features in the decoder. This validates the effectiveness of CLIPUNetr to learn multi-scale information, thus acquiring predictions with more accurate object boundaries and finer structures.

\subsection{Ablation Study}

We validate the effectiveness of each key component used in CLIPUNetr in two parts: architecture ablation and loss ablation. Table \ref{Table_ablation} presents the results.

\begin{table}[h!]
\centering
\scalebox{0.9}{
\begin{tabular}{l c c c c c c} 
 \hline
Dataset & Model & MAE$\downarrow$ & $S_m\uparrow$ & ${wF}_{\beta} \uparrow$ & $maxF_{\beta} \uparrow$ \\ 
 \hline
PhraseCut & CLIPUNetr & \textbf{0.0840} & 0.6975 & \textbf{0.5375} & \textbf{0.5887} \\
 & CLIPSeg-Sides & 0.0844 & \textbf{0.6989} & 0.5326 & 0.5883 \\
 & CLIPSeg-NoSides & 0.1240 & 0.6083 & 0.3779 & 0.4252 \\
 & CLIP-Deconv & 0.0903 & 0.6668 & 0.4683 & 0.5356 \\
 & CLIP-Deconv-BCE & 0.1319 & 0.6468 & 0.3511 & 0.5489 \\
 \hline
UMD+GT & CLIPUNetr & \textbf{0.0025} & \textbf{0.8968} & \textbf{0.7770} & 0.8021 \\
 & CLIPSeg-Sides & 0.0039 & 0.8347 & 0.6707 & 0.6895 \\
 & CLIPSeg-NoSides & 0.0056 & 0.7934 & 0.5450 & 0.5908 \\
 & CLIP-Deconv & 0.0035 & 0.8285 & 0.6694 & 0.6932 \\
 & CLIP-Deconv-BCE & 0.0082 & 0.7772 & 0.4209 & 0.5873 \\
\hline
\end{tabular}
}
\caption{Ablation study on architecture and losses.}
\label{Table_ablation}
\end{table}

\textbf{Architecture Ablation} We take CLIPSeg as the base network, and first extend its decoder to use the same features from layer 3, 6, 9 and 12, denoted as CLIPSeg-NoSides. Then, we extend the decoder with side outputs, denoted as CLIPSeg-Sides. As shown, the model with side outputs achieves superior results.

\textbf{Loss Ablation} We take CLIP-Deconv as the base network, where simple MLP and deconvolution layers are inferred to decode segmentation outputs. CLIP-Deconv is first trained with only binary cross entropy, denoted as CLIP-Deconv-BCE. Then, the training is extended with the hybrid loss, denoted as CLIP-Deconv. In summary, CLIP-Deconv achieves superior results, validating the fact that using hybrid loss is beneficial in learning structure information.

\subsection{Results for Robot Control}
\begin{figure}[h!]
	\centering 
	\includegraphics[width=0.48\textwidth]{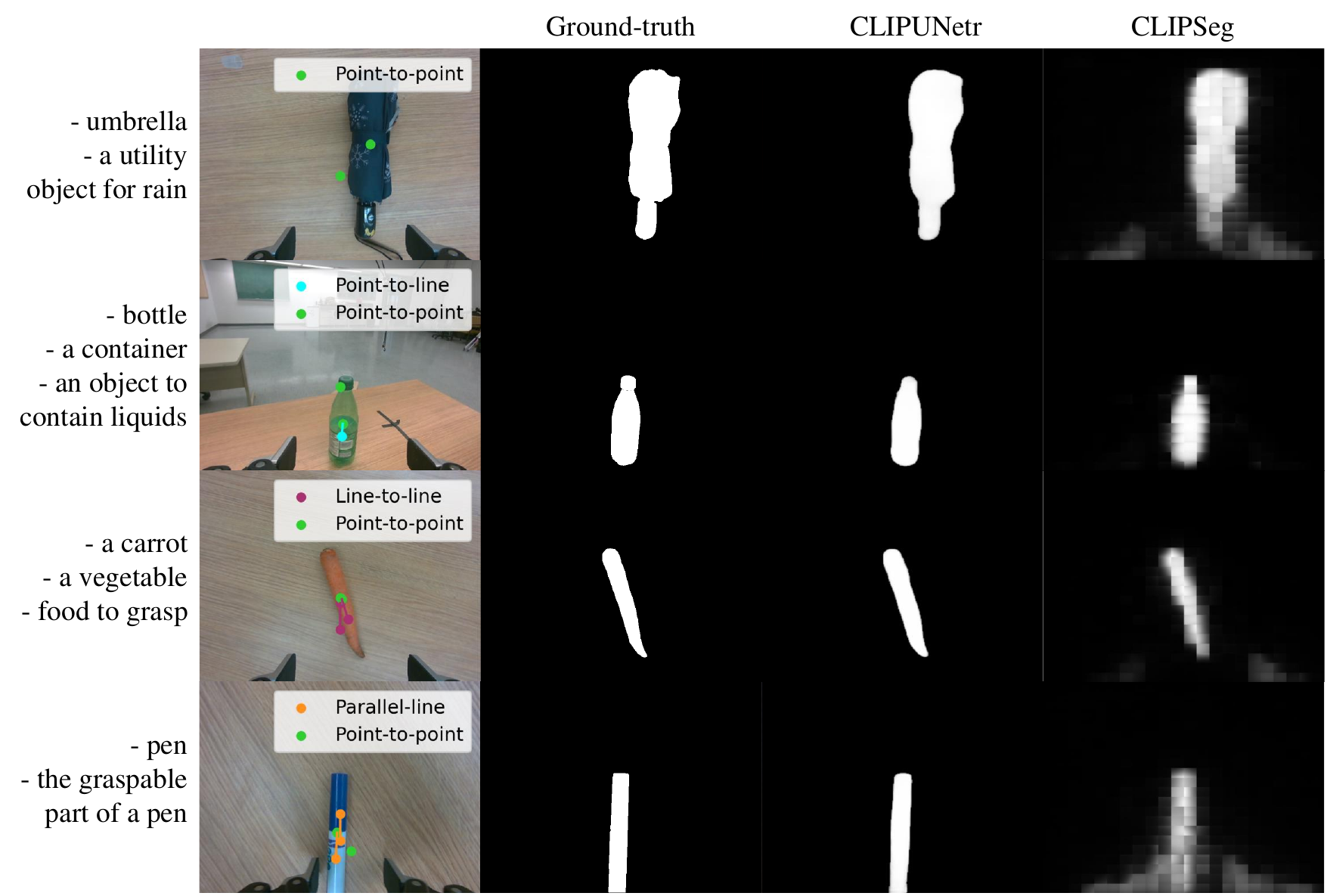}	
	\caption{Quantitative results with robot data.} 
	\label{fig:quantitative_robot}%
\end{figure}

\textbf{Offline Evaluation} Table \ref{Table_Robot} presents the qualitative results of CLIPUNetr with robot data. Our CLIPUNetr achieves superior performance versus CLIPSeg, being able to handle complex motions in unseen robot demonstration videos. Additionally, quantitative comparisons are presented in Figure \ref{fig:quantitative_robot}. Versus CLIPSeg, where predictions are plagued by checkerboard artifacts, CLIPUNetr is able to generate predictions with good quality object boundary and fine structure. This ensures the viability to use CLIPUNetr for robot perception. Additionally, we attach the augmented prompts used to generate the segmentation results. Thanks to the prompt templates used in training, CLIPUNetr is able to better handle diverse prompts containing both affordance and object information. This allows the users to specify prompts in diverse ways, increasing the flexibility of the interface.

\begin{table}[h!]
\centering
\begin{tabular}{l c c c c c c} 
 \hline
Model & MAE$\downarrow$ & $S_m\uparrow$ & ${wF}_{\beta} \uparrow$ & $maxF_{\beta} \uparrow$ \\ 
 \hline
CLIPUNetr & \textbf{0.0101} & \textbf{0.9049} & \textbf{0.8404} & \textbf{0.9104} \\
CLIPSeg \cite{dosovitskiy2020image} & 0.0374 & 0.8075 & 0.4720 & 0.7736 \\
 \hline
\end{tabular}
\caption{Qualitative results with robot data.}
\label{Table_Robot}
\end{table}

\textbf{Online Evaluation} The success rate of robot control over three categories of objects are presented in Table \ref{Table2}. In summary, perception with CLIPUNetr shows comparable performance versus the classical visual interface, being able to successfully segment the unseen daily living objects and perform moving and grasping actions over them. For classical interface, 1 out of 12 tasks fails as a result of poor visual tracking performance. With CLIPUNetr in robot perception, manual clicking can be successfully avoided. And the robot can successfully utilize results of referring expression segmentation to compose geometric constraints and complete the designated manipulation tasks.

\begin{table}[h!]
\centering
\begin{tabular}{l c c c} 
 \hline
Name & Category & Success Rate \\ 
 \hline
w/ CLIPUNetr & Food & 100\%  \\ 
 & Marker Pen & 100\% \\ 
 & Utility & 100\% \\ 
 \hline
Classical \cite{gridseth2016vita} & Food & 80\% \\ 
 & Marker Pen & 100\% \\ 
 & Utility & 100\% \\ 
 \hline
\end{tabular}
\caption{Average success rate of the robot manipulation tasks.}
\label{Table2}
\end{table}


\section{Conclusions}
In this paper, we enhance the human-robot interface of UIBVS with referring expressions. First, we construct CLIPUNetr, where the network adapts CLIP, feature-wise linear modulation and feature scaling with side outputs to perform referring expression segmentation. Second, we build a pipeline to integrate CLIPUNetr into the robot perception and conduct UIBVS control. In comparison, CLIPUNetr improves object boundary and structure measurements by an average of 120\%, producing results with sharper boundaries and finer structures. Integrated into the robot perception, CLIPUNetr can successfully segment objects in an unstructured workspace and assist with UIBVS control. Promising lines of future work include improvements in inference speed, and adaptations of the model in assistive robotics.






{
\bibliographystyle{IEEEtranS}
\bibliography{citation}
}

\end{document}